\documentclass[letterpaper, 10 pt, conference]{ieeeconf}  

\IEEEoverridecommandlockouts                              

\usepackage{graphicx}
\usepackage{booktabs} 
\usepackage{wrapfig}
\usepackage[ruled,vlined]{algorithm2e}

\usepackage{amsmath}

\DeclareMathOperator*{\argmin}{arg\,min}

\newcommand{\Lagr}{\mathcal{L}}

\renewcommand{\baselinestretch}{.992}

\title{\LARGE \bf Visionary: Vision architecture discovery for robot learning}

\author{Iretiayo Akinola$^{1,2*}$, Anelia Angelova$^{1}$, Yao Lu$^{1}$,  Yevgen Chebotar$^{1}$, Dmitry Kalashnikov$^{1}$ \\ Jacob Varley$^{1}$, Julian Ibarz$^{1}$, Michael S. Ryoo$^{1,3}$
\thanks{*Work done during an internship at Google.}
\thanks{$^{1}$Robotics at Google,
        Mountain View, CA and New York, NY, USA.
        }%
\thanks{$^{2}$Columbia University,
       New York, NY, USA
        }%
\thanks{$^{3}$Stony Brook University,
       Stony Brook, NY, USA
        }%
}

\begin{document}

\maketitle
\thispagestyle{empty}
\pagestyle{empty}

\begin{abstract}
We propose a vision-based architecture search algorithm for robot manipulation learning, which discovers interactions between low dimension action inputs and high dimensional visual inputs.
Our approach automatically designs architectures while training on the task -- discovering novel ways of combining and attending image feature representations with \emph{actions} as well as features from previous layers.
The obtained new architectures demonstrate better task success rates, in some cases with a large margin, compared to a recent high performing baseline. Our real robot experiments also confirm that it improves grasping performance by 6\%.
This is the first approach to demonstrate a successful neural architecture search and attention connectivity search for a real-robot task.
\end{abstract}


\section{INTRODUCTION}


For many decades, autonomous robots have been effective in structured environments, such as factories. However, creating robots that work in less-structured environments e.g., households, offices and warehouses, remains challenging and is one of the main goals for autonomous robotics. To operate in diverse and complex environments, autonomous robots have to be intelligent, adaptable, and able to learn from their sensory observations and experience. Recent progress in machine learning research has enabled robotic agents to acquire policies that can map sensory observations, such as images and tactile information, to intelligent actions. For example, large progress has been shown in grasping by learning from raw visual inputs in an end-to-end manner. 

Despite these recent breakthroughs, existing methods still have limitations in terms of adaptability to different environments, task complexity, sample complexity among others. Getting a robot to perform long horizon tasks like packaging in a way that generalizes well to different setting by learning from minimal experiences is still challenging for a number of reasons.
One challenge is designing the visual models which are best suited for performing complex robotic tasks. This is often done by using a network designed for another task, typically borrowing from standard computer vision tasks, such as classification, or by manual design of the architecture.
Furthermore, robot learning happens in a tight vision-action-control feedback loop and the action taken at each step is a crucial input. 
Previous work in robotics have recognized that in addition to feature representation, action representation matters significantly  in robotics manipulation~\cite{finn2015learning}, but is not clear how these two conceptually different sources of data should be combined or connected. 
So far, most learning systems use a simple concatenation of features from all inputs at a single entry point in the network.
It is, however, quite important to consider how the action and other low-dimensional sensory inputs interact and merge with features from high dimensional visual input, as this might affect how well the network captures the interplay between action and the other inputs for accomplishing complex tasks.

\begin{figure}
    \centering
    \includegraphics[width=0.99\linewidth]{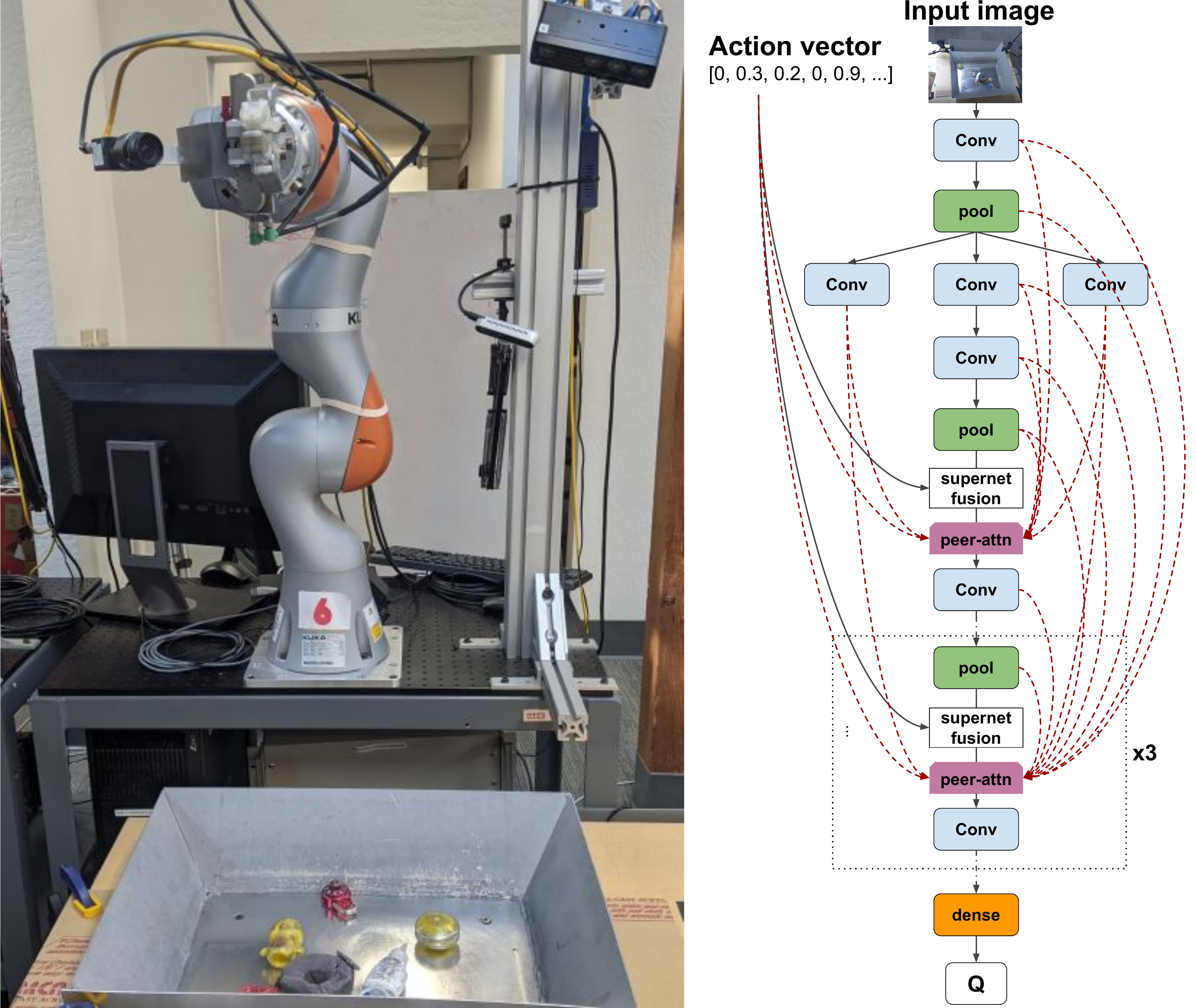}
    \caption{
    We develop a fully differentiable visual architecture search method that can automatically learn a superior architecture while training to adapt to the task on hand. We deploy our approach on a real-robot grasping task.
    }
    \label{fig:robot}
\end{figure}



In this work we propose a novel differentiable architecture search that jointly learns the model architecture while training the model itself for the task at hand. This is done within Reinforcement Learning (RL)-based robot learning context, where in addition to learning the main architecture to generate visual features and combine them with action inputs, the robot is learning a policy to accomplish the task.
Our approach addresses the challenges of learning in the RL setting where the learning is driven by sparse rewards and the robot's own interaction with the environment. RL makes architecture search difficult because the dataset evolves with the skill and experience of the agent during learning; this complex interplay makes it difficult to objectively assess candidate networks and obtain a superior one. Furthermore, iterating over an architecture modification within an RL loop is very expensive. 
The proposed approach alleviates these problems by 
learning to jointly optimize the desired architecture connectivity in one-shot while training the model for the robot manipulation task.

Previous Neural Architecture Search (NAS) approaches \cite{zoph2017neural,zoph2018nas,real2019amoeba} have been successful in visual tasks, e.g. in computer vision for image classification and detection~\cite{liu2018darts}, text analysis~\cite{wang2020textnas} and video~\cite{piergiovanni2018evolving,ryoo2019assemblenet}. However, they have been designed for supervised problems only, where a direct label is available per example. Secondly, they are targeting specific vision tasks, and do not consider the vision-control loop with action input modality.
Furthermore, while successful, NAS approaches are often obtained at a very high computational cost~\cite{strubell2019energy}. 
Combining these with the computationally intensive exploration needed by a sparse-reward RL algorithms, results in an infeasible exploration task.
Our approach specifically targets RL-based robot learning, exploring the combination of action and visual inputs quickly and efficiently via a differentiable one-shot architecture search.

Our proposed algorithm, which we term \textit{Visionary},  leads to the discovery of new and successful architectures (Figure~\ref{fig:models}). They feature novel components 
which 1) introduce learned combinations of robot actions and visual inputs,
2) are highly suited to robotics tasks i.e. directly optimized for such tasks, and 3) are applicable to different tasks including real-robot grasping (Figure~\ref{fig:robot}).
Our contributions are as follows:
\begin{itemize}
    \item A novel approach to automatically design the robot vision system for closed-loop vision action control for a grasping task. The architecture search is differentiable and learns to combine vision and action inputs while simultaneously training the network.
   
    
    \item  Our architecture search addresses the challenges related to sparse-rewards exploration and computational feasibility in robotics. It is general and flexible and applicable to tasks requiring vision and action inputs.


    \item We demonstrate that model architecture search yields performance gains on a real-robot task. 
    In particular, we confirm the effectiveness of \emph{peer-attention} to learn attention connectivity between visual inputs and actions inputs, discovering architectures for the task at hand that outperform previous known architectures.
\end{itemize}

\section{Related work}

\textbf{Vision-based robot learning.} Vision-based robot learning has a long history in robotics. Prior to recent breakthroughs in deep learning, hand-crafted features were extracted from images for robot learning~\cite{nguyen2014autonomously}. More recently, deep learning has become the de-facto approach for learning from images after achieving superior performance on different computer vision problems. Robotics has since leveraged findings in computer vision by successfully applying deep learning-based  perception module in high performing methods to solve different robot tasks such as grasping~\cite{gualtieri2015high,lenz2015deep,redmon2015realtime,cadena2016multi,mahler2017dexnet,xiang2018posecnn,berscheid2019improving,lee2019icra,xie2019thebest}.   
End-to-end deep learning systems for vision-based robot learning, for example~\cite{pinto2016supersizing,levine2016e2e,levine2016learning,viereck2017learning,mahler2017dexnet,byravan2018se3pose,pathak2018zero,zeng2018robotic,kalashnikov2018qtopt,quillen2018deep}, have demonstrated early success of fully learnable systems.
\cite{viereck2017learning} proposed to use an end-effector mounted depth sensor for easier sim2real transfer to assist in learning a closed loop controller for grasping. 
~\cite{danielczuk2019mechanical,danielczuk2020xray} proposed a general object search and grasping formulation, `Mechanical Search', which consists in a two-stage perception and search policy pipeline.
QT-Opt of~\cite{kalashnikov2018qtopt} demonstrated an end-to-end RL approach which learns from raw pixels and is able to acquire sophisticated grasping skills. 
Unlike preceding work, we demonstrate one can fully adapt the learning process building the most suitable architectures and learning the grasping skills concurrently with acquiring the skills from data.

	
\textbf{(Visual) Architectures for robot learning:}	
Earlier deep learning architectures for grasp perception~\cite{levine2016e2e} explicitly learned a fixed number of local features in the image plane which are then merged with the action vectors.
Subsequently, more general approaches have been established which 
combine the learned visual embeddings with the action vectors~\cite{wu2020learning,levine2016learning,viereck2017learning,kalashnikov2018qtopt}.
These works typically obtain a standard deep neural net, often adapted from computer vision tasks, 
and combine the action vector in an adhoc manner e.g. by concatenation with the visual features~\cite{viereck2017learning,levine2016learning,kalashnikov2018qtopt}. 
Our approach provides a direct exploration of more optimal functional transformations and combinations with the action inputs, as guided by the robot learning task at hand.

\textbf{Neural Architecture Search.}
Broadly speaking, there are three classes of neural architecture search methods.
Some methods~\cite{miikkulainen2019evolving,liu2017hierarchical} use evolutionary algorithms to rank, mutate and evolve candidate networks to produce better performing ones. The second class takes a differentiable approach to learn and prune weights of a hybrid network which is a union of candidate network model combined by learned weights~\cite{liu2018darts}. The third class uses RL to train a network-generating agent that generates model description to maximize reward function~\cite{zoph2016neural}. A survey ~\cite{kyriakides2020introduction} discusses the pros and cons of each class. Our work uses a differentiable approach to obtain a high performing Q-learning model in the RL setting for robotic manipulation tasks. 

\begin{figure*}
    \vspace{0.3cm}
    \centering
    \includegraphics[width=0.7\linewidth]{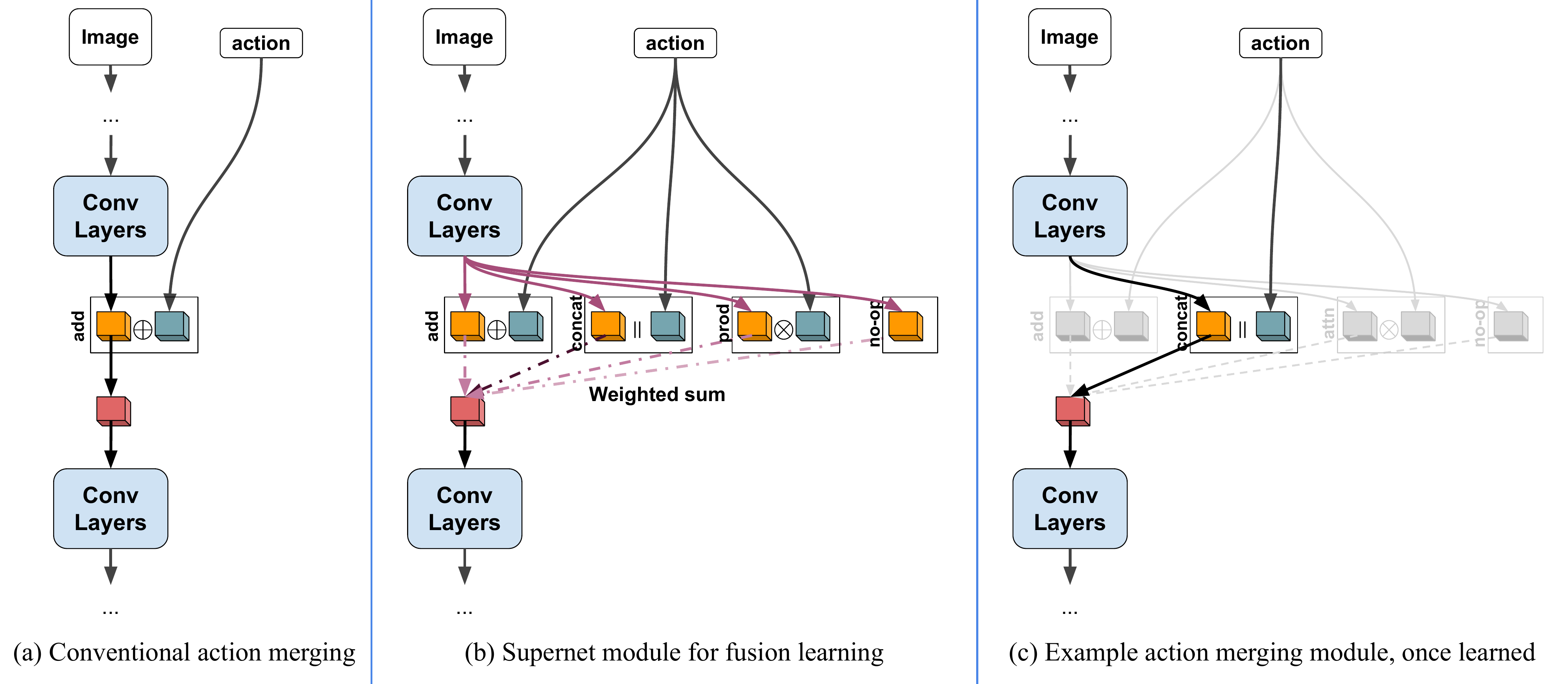}
    \caption{Comparison between (a) the conventional action merging module that uses the addition operation versus (b) our proposed supernet fusion module that is a weighted combination of a few fusion operations. 3D cuboids indicate state representations, which are being fused. Dotted lines are for weighted connections. Unlike prior CNN models that rely on hand-designed state-action fusion, our supernet module allows discovering better merging operations by learning their weights. Once learned, we only maintain the merging operation with the largest weight (c), making the model computationally efficient.
    }
    \label{fig:fusion}
    \vspace{-0.5cm}
\end{figure*}


\section{Preliminaries}
\label{sec:preliminaries}

\textbf{Reinforcement Learning (RL) Formulation:} Using the RL framework, we formulate robot manipulation as a Markov Decision Process (MDP) where the robot interacts with the environment to maximize the expected reward. The agent makes observations $s \in \mathcal{S}$ in the environment, takes action $a \in \mathcal{A}$ and gets rewards according to a reward function $\mathcal{R}(s, a)$ that captures the desired behavior. The goal of RL is to maximize the expected discounted cumulative reward.

We will use a variant of Q-Learning algorithm for continuous actions (QT-Opt~\cite{kalashnikov2018qtopt}) to learn state-action value function, also referred to as the Q-value function. This Q-value function, given as $\mathcal{Q}^{\pi}(s,a) = r(s,a) + \gamma  \max_{a'}\mathcal{Q}^{\pi}(s',a')$, captures the discounted sum of rewards that the agent can accumulate starting from a state $s$, executing action $a$ and following the policy $\pi$. $s'$ is the next state and $\gamma$ is a discount coefficient which trades off current with future rewards. 
The policy $\pi(s)$ that gives the action for each state is obtained by optimizing the learned Q-function using the cross-entropy method-- a derivative-free optimization algorithm (see~\cite{kalashnikov2018qtopt}). 

The focus of this work is on discovering superior neural architectures that produce more expressive Q-functions. We observe that the robot's observation can be multimodal, i.e. consists of components that have different dimensions, and introduce an algorithm to discover better ways to wire up neural networks for multi-modal Q-functions for the task.


\textbf{Robot Setup (Simulation \& Real)}
Our experimentation platform for learning is adapted from QT-Opt~\cite{kalashnikov2018qtopt}.
We develop and test our approach in simulation (using the PyBullet-based physics simulator \cite{coumans2016pybullet}). 
We validate our results on the grasping task using real hardware.



All the tasks have the same state space and action space:

\textbf{Inputs.} We use an RGB image of size 472x472 from a static camera mounted above the workspace over the shoulder of the robot arm.
In addition to the image, the robot state includes a binary indicator of whether the gripper is opened or closed (gripper-state), the height of the gripper above the bottom of the tray (a scalar). The full observation is be defined as $s_t = (I, gripper\_state, height)$.

\textbf{Actions.} The action space consists of a gripper pose displacement, and an open/close command. The gripper pose displacement is a difference between the current pose and the desired pose in Cartesian space, encoded as translation $T \in R^3$ , and vertical rotation encoded via a sine-cosine encoding $R \in R^2$. A gripper command (whether open or closed) is encoded as one-hot vector $[gripper\_command] \in {0, 1}$. The full action is defined as $a_t = (T, R, gripper\_command)$.

\textbf{Reward function.} 
We use a sparse reward where the agent obtains a reward of 1.0 if successful on the manipulation task and tiny penalty (-0.01) at each step of the episode.
The success of a manipulation task is task-specific. For example, for grasping it means that the object is suspended by the gripper above the bin. 
For stacking, a success is declared when one block is stably placed on top of another. 

\textbf{Baseline neural architecture.} As a strong baseline we use the visual architecture described in the QT-Opt~\cite{kalashnikov2018qtopt} algorithm that accomplished impressive grasping performance. 
It is primarily a non-recurrent neural network which integrates the action vector after the fifth convolutional layer. The action merging is done by addition (Figure~\ref{fig:models} (c)).


\section{Approach}
\label{sec:main}


\subsection{Robot architecture learning formulation}
\label{subsec:formulation}

In addition to learning the weights of a Q-Value function for vision-based control, we learn the neural architecture itself to accomplish the task more successfully. 
Specifically, we formulate a one-shot architecture search method that performs joint optimization for the final task.



Let 
$Q(s, a)$ be the Q-Value function that takes state-action as an input (image and action vector) and produces a Q-Value.
We can think of $Q(s, a)$ as dependent on two sets of learnable parameters: $Q(s, a) = Q_{\phi}(s, a ; \theta)$, where $\phi$ is the parameters governing the architecture of the CNN function (e.g., what layer to use, how layers are connected, etc.), and $\theta$ is the set of parameters of the corresponding CNN (e.g., convolution filter values).
The objective to find high performing neural architectures for robot learning can be formulated as a bi-level optimization (Eq.\ref{eq:two-optimization}), where one maximizes the Q-Value for a large number of sampled architectures, each of which need to be trained on the task:
\begin{equation}
\begin{split}
    \phi^* &= \argmin_{\phi} \Lagr_{val}(Q_{\phi}(s, a ; \theta^*))\\
    \text{s.t.}~~ \theta^* &= \argmin_{\theta} \Lagr_{train}(Q_{\phi}(s, a ; \theta))
\end{split}
\label{eq:two-optimization}
\end{equation}
$\Lagr(\cdot)$ is a Bellman loss described in QT-Opt~\cite{kalashnikov2018qtopt}.
However, directly optimizing the outer loop of Eq.\ref{eq:two-optimization} is very expensive and challenging, especially for large number of parameters in  $\phi$, as each step requires the entire inner loop optimization. 
Further, the typical blackbox optimization strategy (e.g., \cite{real2019amoeba}) for this problem suffers from randomness in the inner loop RL training and evaluation, as it requires a reliable `fitness' score to be measured for each architecture.

Instead of tackling this problem with the bi-level optimization, we take an alternative strategy: one-shot differentiable joint optimization. That is, we jointly optimize the parameters in the original outer loop and the parameters in the original inner loop, in a single loop (Eq.~\ref{eq:one-optimization}). This is possible when $\phi$ is formulated to be differentiable~\cite{liu2019darts} with respect to the training loss, allowing the gradients to flow end-to-end. 
\begin{equation}
\begin{split}
    \phi^*, \theta^* &= \argmin_{\phi,\theta} \Lagr_{train}(Q_{\phi}(s, a ; \theta))
\end{split}
\label{eq:one-optimization}
\end{equation}


Our differentiable architecture discovery formulation allows directly regressing optimal architecture parameters jointly with the CNN filter values during the RL training. This enables plugging in our approach into any existing reinforcement learning framework, simply by replacing the previous model with our one-shot discovery model.




\subsection{Architecture search space}
\label{sec:arch}




An important component of architecture search is defining the parameters of the search. Most vision-based robotic manipulation systems fuse information from high dimensional image inputs with low dimensional vectors such as actions or proprioceptive information. There are different design choices to combine these multi-modal inputs. Herein lies the crux of our approach. We build the dimensions of the search around the possible merging approaches for combining these observations.


\textbf{Action Merging:}
We first consider merging in the robot action into the full visual architecture. The Q function in general takes two different types of inputs: (1) a visual state representation which usually has a form of a high-dimensional tensor with spatial information, and (2) an action representation which often is a 1-D vector. In order for the Q function learning to be reliable, such two representations, heterogeneous in shape and semantics, are required to be optimally merged.
In this paper, we focus on understanding two key technical questions: (i) how should we merge the action representation with the other representations (i.e., by what operations) and (ii) where should we merge them (i.e., locations).

\paragraph{Action merging operations}
A number of linear and non-linear representation merging methods are considered. Let $x_i$ be the visual state representation at the level $i$ and $a$ be the action representation. 
The action merging operations the architecture search considers include:
\begin{enumerate}
\item Addition: $x_{i+1} = x_i + T(a)$ 
\item Concatenation: $x_{i+1} = x_i \mathbin\Vert T(a)$
\item Hadamard product: $x_{i+1} = x_i \odot T(g(a))$ 
\item Action to convolutional filters: $x_{i+1} = x_i * c(a)$ 
\end{enumerate}
where $T(\cdot)$ is a tensor shape expansion function to match the shape of $x$ and $a$.
$g(\cdot)$ is a fully connected layer generating an output whose size is identical to the spatial shape of $x$ and $c(\cdot)$ is a fully connected layer whose output is reshaped to form the filters for the convolution operation $*$. 

\paragraph{Action merging locations}
We are able to apply any of the above operations to any intermediate visual state representation $x_i$ at the location $i$, producing the next intermediate representation $x_{i+1}$. This could be done once or multiple times, and the location $i$ could be at the early stage of the CNN architecture or at the later stage. 



\begin{figure*}
    \vspace{0.1cm}
    \centering
    \includegraphics[width=0.77\linewidth]{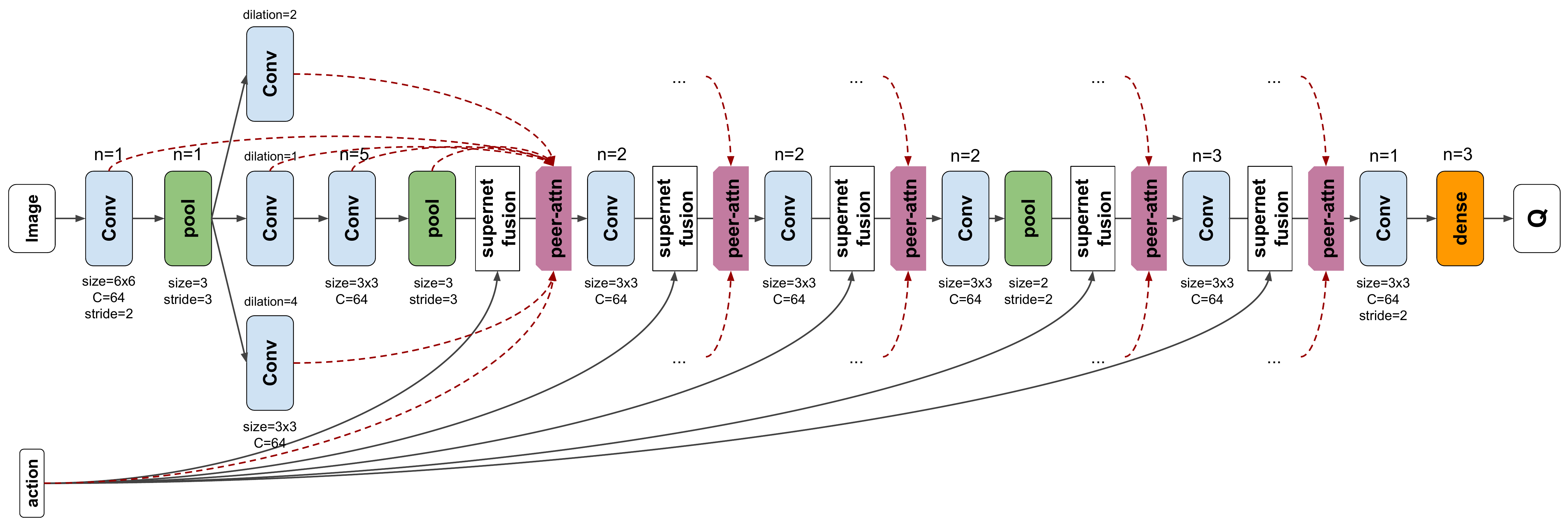}
    \caption{The final model for our one-shot architecture discovery. The model is fully differentiable, and training it with RL allows the discovery of the CNN architecture optimized for the policy. Once the architecture is learned, it uses a comparable number of parameters to the baseline.
    }
    \label{fig:full}
    \vspace{-0.2cm}
\end{figure*}

\textbf{Attention Connectivity:}
We also search for complex connectivity among visual representations and action representations, modeled in terms of an attention mechanism. Specifically, we take advantage of \emph{peer-attention} and consider connection between every pair of layers in the CNN architecture. This explicitly enables visual representations to be conditioned based on action representation via attention, which we discuss more in Section \ref{subsec:attention}. The dotted red lines in Figures \ref{fig:robot}, \ref{fig:full}, and \ref{fig:models} illustrate such attention connections.


\begin{figure}
    \vspace{0.3cm}
    \centering
    \includegraphics[width=0.85\linewidth]{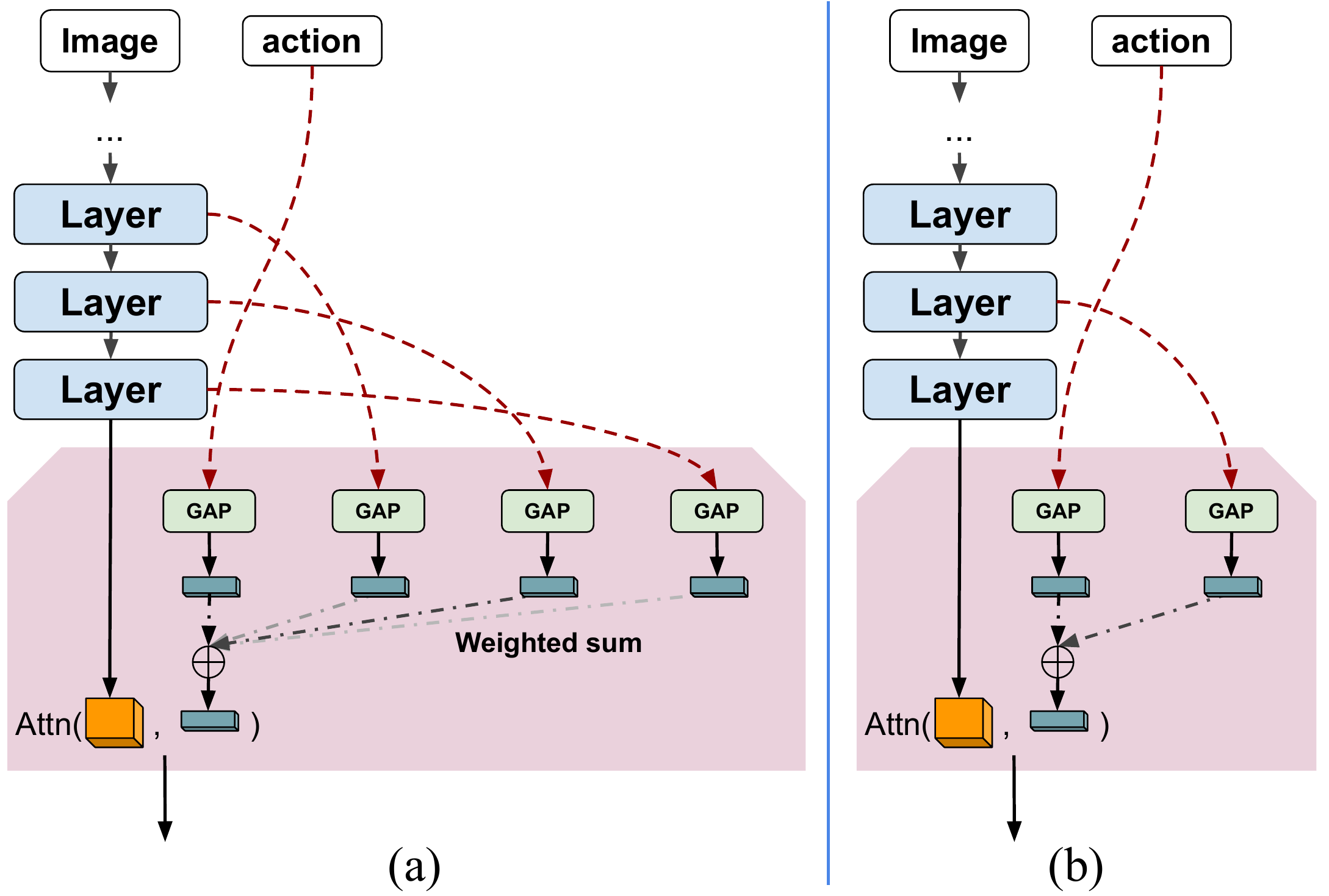}
    \caption{(a) An illustration of the peer-attention module during its learning. (b) An example showing how a peer-attention module may converge after learning. GAP stands for `global average pooling'. 
    }
    \label{fig:peer-attention}
    \vspace{-0.35cm}
\end{figure}

\subsection{Differentiable action merging search}
\label{subsec:differentiable}

Instead of non-differentiable architecture exploration by discretely adding potential components iteratively, we include all possible action-merging operations after every computational layer, excluding the very early layers, 
to the computation graph using a `supernet' module (as shown in Figure~\ref{fig:fusion}). The learnable weights ultimately determine which operations are important to obtain a high-performing model.

The supernet's action-merging 
(i.e., fusion) module consists of a number of action merging blocks $f_j$, each of which takes in an intermediate representation tensor $x$ and the action vector $a$ as an input. The output of the module is given as: $x_{i+1}=f_j(x_i, a)$ where $x_i$ is the intermediate representation at the $i$th layer of the CNN model while each action merging operation $f_j$ is applied and combined to obtain a new representation $x_{i+1}$. For instance, if $f_1$ is addition and $f_2$ is concat, $f_1(x,a)= x+T(a)$ and $f_2(x,a)= x || T(a)$. We learn a set of weights $w_{ij}$ (at each level $i$), corresponding to each candidate operation $f_j$. 
The output of the supernet module $F_i$, given the input $x$ and $a$ is computed as:
\begin{equation}
F_i(x,a) = \sum\nolimits_j \sigma(w_{ij}) \cdot f_j(x, a)
\end{equation}
where $\sigma(w_{ij})$ is formulated with a softmax function: $\sigma(w_{ij}) = e^{w_{ij}} / \sum_k e^{w_{ki}}$
The supernet module computes a weighted summation of all possible operations while constraining the weights with softmax, thereby making a soft-selection of one of the operations. Once these weights are learned and finalized, non-argmax connections are pruned, making it spend the same amount of compute to conventional action merging during the inference.


We note that we have included a no-op possibility (i.e., $f(x,a)= x$) to not use any action merging. 
This allows placing our supernet module at every potential action merging locations within the CNN, and let the learning decide which merging locations to use based on the training data. 


\subsection{Differentiable attention connectivity search}
\label{subsec:attention}


Channel-wise attention, re-weighting channels differently conditioned by another representation, has been very successful in many perception tasks such as object recognition \cite{hu2018squeeze} and video classification. In particular, the concept of peer-attention~\cite{ryoo2020assemblenetplus} allows efficient modeling of layer connectivity using an attention mechanism. In order to make our layers capture the relations between the action representation and the image-based state representations, we search for the peer-attention connection as a part of our architecture discovery.




We formulate our peer-attention module similar to the supernet module, so that the optimal connection is found through the gradient computation of differentiable parameters. Let the function $G_i(x_i)$ denote the peer-attention module taking the input $x_i$ at location $i$ within the network
\begin{equation}
G_i(x_i) = \sum_{v \in \{a, x_1, \cdots, x_i\}}  \text{Attn}\big(x_i, ~h_{iv} \cdot \text{GAP}(v)\big)
\end{equation}
where $h_{iv}$ is a scalar weight corresponding to the representation $v$ (constrained with sigmoid), $\text{GAP}(\cdot)$ is a (spatial) global average pooling layer, and $\text{Attn}(\cdot)$ is the attention function often implemented as a broadcasted element-wise multiplication. Figure~\ref{fig:peer-attention}) illustrates the process. The use of $\text{GAP}(\cdot)$ makes the entire module to be performing channel-wise attention, which is different from the Hadamard product used as one of the action merge operations described in the previous subsection.

Importantly, the representation $v$ our peer-attention considers includes the action representation $a$. This allows any particular representation $x_i$ to be potentially conditioned by the action $a$ (whose strength is learned with the differentiable parameter $h_{iv}$). Also notice that maintaining multiple peer-attention connections (and fusing them) is very computationally inexpensive, as it essentially is a weighted summation of 1-D vectors with only channel dimension, without any spatial resolution. Peer attention adds 0.35\% increase
in the total number of FLOPS.

\subsection{Full one-shot architecture search}
\label{subsec:full_model}

We focus on off-policy RL as the test-bed for our search as it enables us to learn from fixed datasets which is necessary for fair comparison of different architectures during the search. Specifically, instead of using the exploration strategy defined in the RL algorithm, we collect and set aside a fixed dataset of experience. The experience episodes are continuously sampled from a replay buffer, letting the downstream network learn from the same offline data.

\begin{figure}
    \centering
    \includegraphics[width=0.99\linewidth]{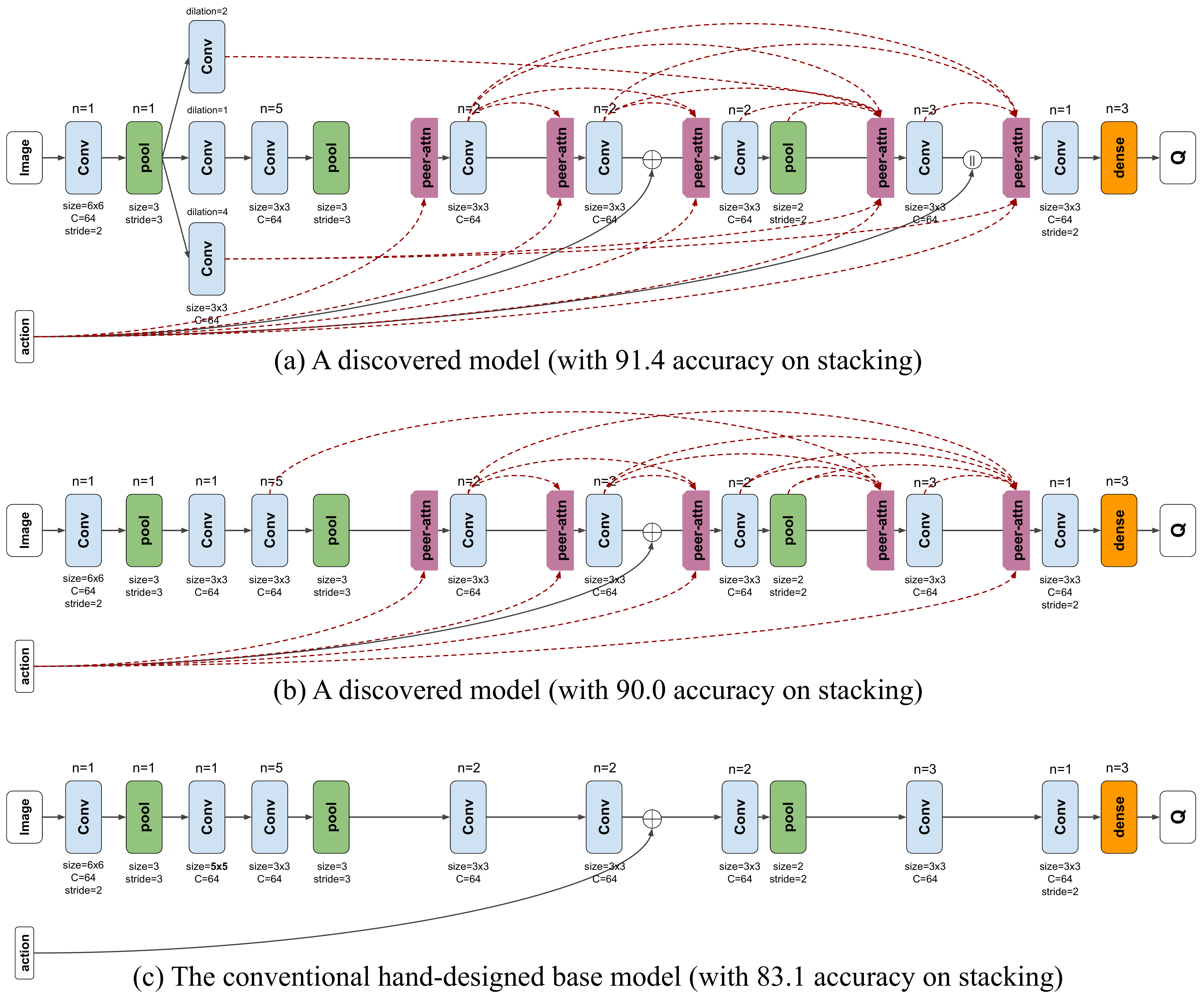}
    \caption{Comparing discovered architectures with the conventional baseline model. Their computation time is similar thanks to efficient peer-attentions. We show peer-attention connections with weights greater than 0.5.
    }
    \label{fig:models}
    \vspace{-0.35cm}
\end{figure}

Figure~\ref{fig:full} shows the final model for our one-shot differentiable architecture, using both the supernet modules and the peer attention modules. Once the architecture is found (by training it once), the result becomes the final fixed architecture (e.g., Figure~\ref{fig:models} a, b).  
Compared to the conventional robot learning architectures having one action addition merging in the middle (Figure~\ref{fig:models} c), our one-shot architecture discovery model shows three differences: (1) We added supernet modules at five potential action merging locations. (2) We added a peer attention module after every supernet module. (3) In order for the peer-attention modules to take advantage of more diverse intermediate representations, we added two more parallel convolutional layers with spatial dilation rate of 2 and 4. Note that the outputs of these parallel layers are only connected through peer-attention modules and are not directly fused with other representations.

The full search model requires 41.09 GFLOPS, while the found model (Figure~\ref{fig:models} a) uses 41.05 GFLOPS. They are comparable to the base model (Figure~\ref{fig:models} c) requiring 40.91.

\section{Experimental Results}
\label{sec:results}
We evaluate our approach on tasks in simulation and on a physical robot. 
Our one-shot search model enables its architecture discovery to be done by training it once with the training data. This is done by directly replacing the conventional CNN model with our full search model 
within the standard RL framework. The parameters deciding the final architecture structure 
are updated through the standard backpropagation, computing the gradient based on the loss of the RL algorithm.
We use the same hyper-parameters for the architecture search and the final training process.
Specifically, our batch size is set to be 4096. All weights are initialized with truncated normal random variables and L2-regularized with a coefficient of 9e-5. Models are trained with SGD with momentum, using learning rate 0.0044 and momentum 0.958.
We train the models for 80K training iterations for sim, and $\sim$110K iterations for the real robot.


\begin{table}
\vspace{0.2cm}
\caption{Performance evaluation on the block stacking task. We measure the task success rate (in \%), computing their mean and standard deviation (indicated with $\pm$). 
We report both the mean and the highest. The superiority of Visionary is statistically significant with the p-value less than 0.001.}
\centering
\begin{tabular}{lcc}
\toprule
Model & Mean success rate & Best success rate \\
\midrule
Base model (from \cite{kalashnikov2018qtopt}) & 74.8 $\pm$ 7.3 & 83.1 \\
Visionary (ours) &\textbf{84.1} $\pm$ 4.4 & \textbf{91.4} \\
\bottomrule
\end{tabular}
\label{tab:main}
\vspace{-0.4cm}
\end{table}

\begin{figure}
    \centering
    \includegraphics[width=0.75\linewidth]{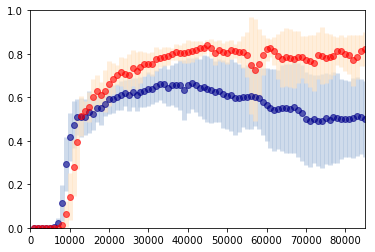}
    \vspace{-0.25cm}
    \caption{Mean and standard deviation of our model (red) compared to the conventional baseline (blue) at every 1K time steps. The X-axis is the training iterations, and the Y-axis is the task success rate. 
    }
    \label{fig:task37}
    \vspace{-0.2cm}
\end{figure}

\subsection{Quantitative performance evaluation in simulation}

Table~\ref{tab:main} shows performance of the proposed approach compared to the baseline (from  \cite{kalashnikov2018qtopt, akinolad2020learning}) on the block stacking task in simulation. 
Since the RL algorithms exhibit a lot of variance, we run the experiments for 15 times for each Visionary and the baseline. For each run, at every 1K training steps, the evaluation was done with 200 robot trials 
of the stacking task with random object locations.
As seen from Table~\ref{tab:main}, the proposed architecture outperforms prior work by a large jump in task success rate. 
Figure~\ref{fig:task37} illustrates the performance differences during training.

\textbf{Significance test.} We ran the standard independent two-sample Student's t-test for measuring the statistical significance of the Visionary performance compared to the baseline from Table~\ref{tab:main}. The p-value is less than 0.001 which suggests that it is very statistically significant (t is 4.23 and df is 28).


\textbf{Ablation experiments} reveal that the performance gains of our approach are due to the combination of better action merging and the learned layer connectivity (Table~\ref{tab:ablation}). 
We confirm that various components of our approach benefit the architecture discovery. 
We compare our approach against strong prior architecture search based on an evolutionary algorithm, and it is not as effective partially due to the variance in RL training. Our one-shot differentiable search model is more robust, as it is directly gradient based.

\begin{table}
\vspace{0.2cm}
\caption{Ablation experiments. Stacking task. Task success rate (\%).}
\centering
\begin{tabular}{lc}
\toprule
Method & Success rate \\
\midrule
Base model  & 74.8 \\
Evolutionary algorithm (used in \cite{real2019amoeba,piergiovanni2019tvn}) & 74.8 \\
Supernet module search only & 75.1 \\
Peer-attention search only & 83.0 \\
Full  & \textbf{84.1} \\
\bottomrule
\end{tabular}
\label{tab:ablation}
\vspace{-0.2cm}
\end{table}

\begin{table}
\caption{Real-Robot Grasping Task Performance. We measure the task success rate (\%), over 810 robot task trials.}
\centering
\begin{tabular}{lcc}
\toprule
Model & Success rate\\ 
\midrule
Base model (from ~\cite{kalashnikov2018qtopt}) & 64.6 \\ 
Visionary (ours) &\textbf{70.8}\\
\bottomrule
\end{tabular}
\label{tab:real}
\vspace{-0.3cm}
\end{table}

\subsection{On-Robot Grasping Experiments}
\label{sec:real-robot}

\begin{figure}
    \centering
    \includegraphics[width=0.28\linewidth]{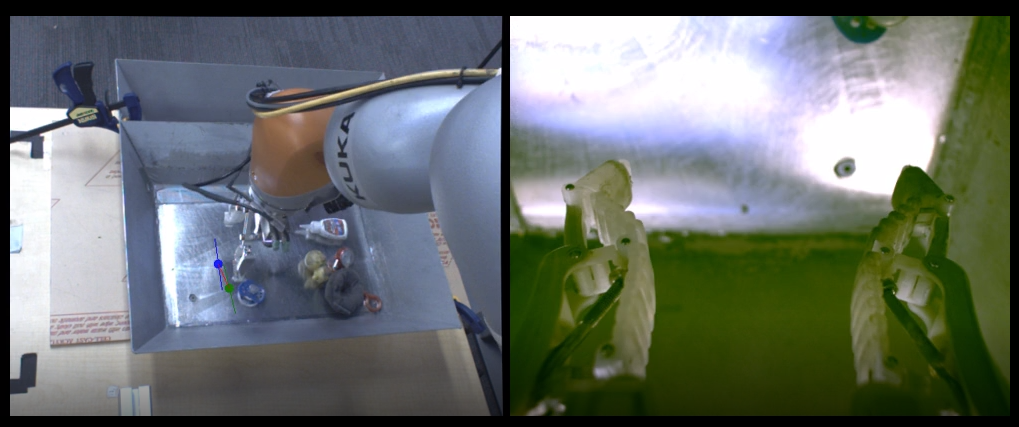}
    \includegraphics[width=0.28\linewidth]{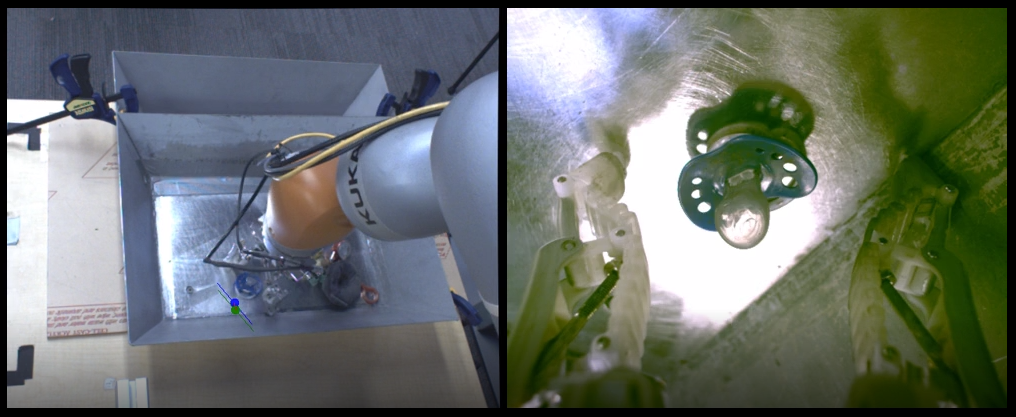}
    \includegraphics[width=0.28\linewidth]{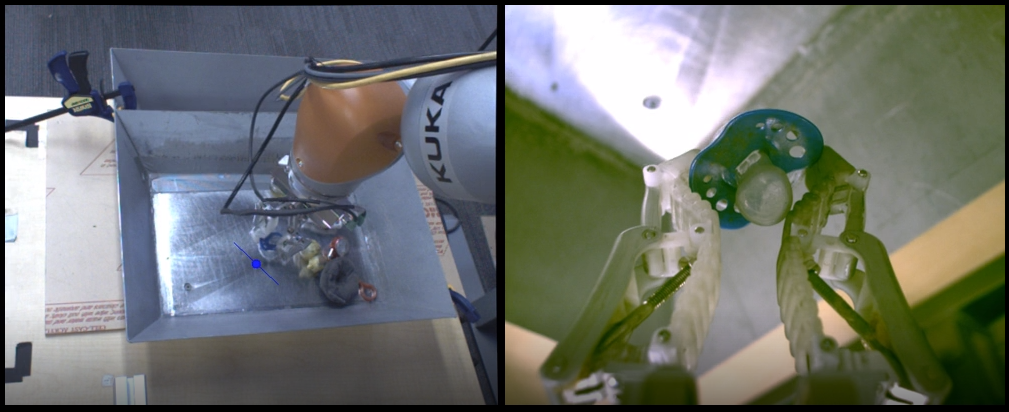}
    \vspace{0.05cm}
    
    \includegraphics[width=0.28\linewidth]{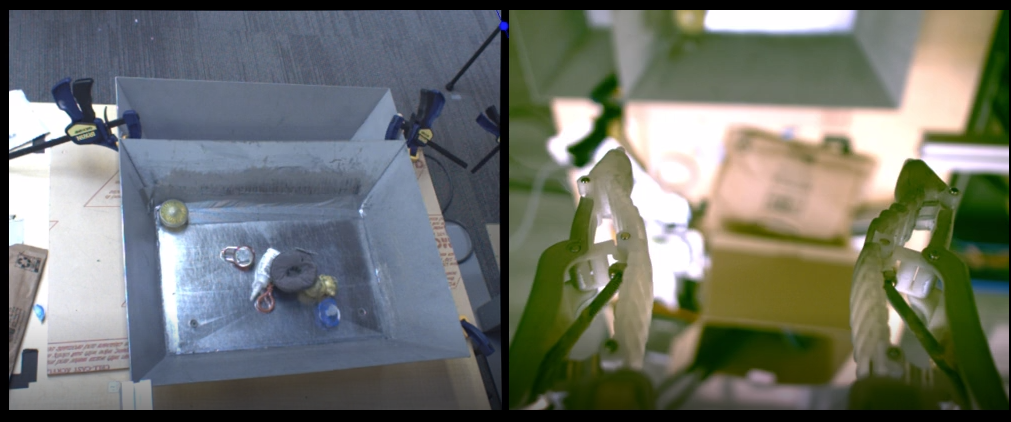}
    \includegraphics[width=0.28\linewidth]{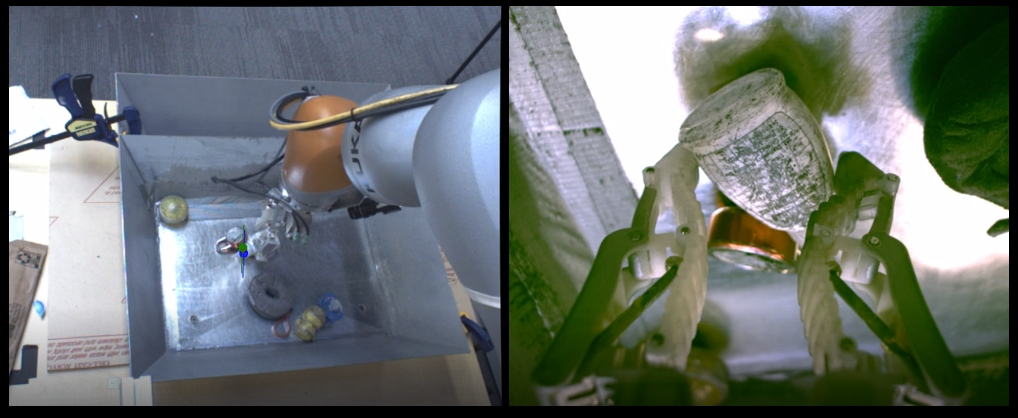}
    \includegraphics[width=0.28\linewidth]{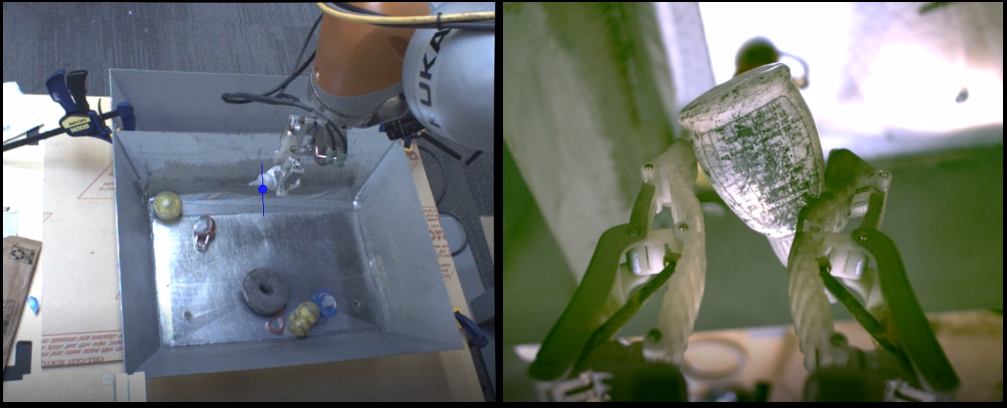}
    \vspace{-0.1cm}
    \caption{On-robot experiments. Grasping task. KUKA robot arm. 
    Left: Beginning of the episode, Middle: approaching an object to grasp (top: a pacifier; bottom-a small bottle). Right: Object successfully grasped. 
    }
    \label{fig:kuka-robot}
\end{figure}

We further test our approach on a KUKA Robot for the task of grasping random objects from a bin (Figure~\ref{fig:robot}). Using offline real robot data \cite{levine2020offline}, the model architecture is discovered and learned. It is then tested on the real-time KUKA robot in a different environment, which is more 
challenging and mimics setups in a warehouse or a factory floor. 
%
Table~\ref{tab:real} shows the grasping task success rates on the real robot. 
We observe that the proposed models perform better than the baseline on real robot grasping tasks.
Figure~\ref{fig:kuka-robot} shows snapshots from the on-robot experiments. 

\section{Conclusion}
\label{sec:conclusion}
We present a novel approach for differentiable architecture search for robot learning with vision and action inputs, which simultaneously develops the architecture while traininng it for the final robot task. We obtain meaningful improvements in simulation and on real-robot tasks.  


{\small
\bibliographystyle{IEEEtran}
\bibliography{example}
}

\end{document}